\documentclass[conference]{IEEEtran}
\IEEEoverridecommandlockouts
\usepackage{cite}
\usepackage{dsfont}
\usepackage{bm} 
\usepackage{url}
\usepackage{amsmath,amssymb,amsfonts}
\usepackage{algorithmic}
\usepackage{graphicx}
\usepackage{textcomp}
\usepackage{xcolor}
\usepackage{subfigure}
\usepackage{booktabs}
\usepackage{setspace}
\usepackage{hyperref}
\def\BibTeX{{\rm B\kern-.05em{\sc i\kern-.025em b}\kern-.08em
    T\kern-.1667em\lower.7ex\hbox{E}\kern-.125emX}}

\begin{document}

\title{End-to-End Modeling of Hierarchical Time Series Using Autoregressive Transformer and Conditional Normalizing Flow-based Reconciliation}

\author{
\IEEEauthorblockN{Shiyu Wang, Fan Zhou, Yinbo Sun, Lintao Ma, James Zhang, Yangfei Zheng}
\IEEEauthorblockA{\textit{Ant Group} \\
Hangzhou, China \\
\{weiming.wsy, hanlian.zf, yinbo.syb, lintao.mlt, james.z, yangfei.zyf\}@antgroup.com
}
}

\maketitle

\begin{abstract}
Multivariate time series forecasting with hierarchical structure is pervasive in real-world applications, demanding not only predicting each level of the hierarchy, but also reconciling all forecasts to ensure coherency, i.e., the forecasts should satisfy the hierarchical aggregation constraints. Moreover, the disparities of statistical characteristics between levels can be huge, worsened by non-Gaussian distributions and non-linear correlations. To this extent, we propose a novel end-to-end hierarchical time series forecasting model, based on conditioned normalizing flow-based autoregressive transformer reconciliation, to represent complex data distribution while simultaneously reconciling the forecasts to ensure coherency. Unlike other state-of-the-art methods, we achieve the forecasting and  reconciliation simultaneously without requiring any explicit post-processing step.  In addition, by harnessing the power of deep model, we do not rely on any assumption such as unbiased estimates or Gaussian distribution. Our evaluation experiments are conducted on four real-world hierarchical datasets from different industrial domains (three public ones and a dataset from the application servers of Alipay \footnote{Alipay is the world’s leading company in payment technology.  https://en.wikipedia.org/wiki/Alipay }) and the preliminary results demonstrate efficacy of our proposed method.
\end{abstract}

\begin{IEEEkeywords}
hierarchical time series, transformer, conditioned normalizing flow, reconciliation\end{IEEEkeywords}

\section{Introduction}
Many real-world applications \cite{athanasopoulos2009hierarchical,jeon2018reconciliation,dangerfield1992top} involve simultaneously forecasting multiple time series that are hierarchically related via aggregation operations, e.g., department sales of multiple stores at different locations and traffic flow in hierarchical regions. These time series not only interact with each other in the hierarchy, but also imply coherency, i.e., time series at upper levels are the aggregation/summation of those at lower levels.  For instance, as shown in the hierarchical structure of Australian domestic tourism demand \cite{bushell2001tourism}, the data contains 4 levels from top to bottom including, 1 country, 7 states, 27 zones, and 82 regions, among which, different levels of forecasts have distinct goals.  Specifically, bottom-level forecasts are often about specific demand to help with regional government decisions, while upper-level forecasts look at the macro perspective to assist national strategies. On top of the general statistical disparities between levels in real-world hierarchies, the entanglement of their interactions and correlations presents a great challenge to the prediction model.  Therefore, how to exploit the information among hierarchical time series, while learning from non-Gaussian data distributions and non-linear correlations, becomes a critical task for improving the prediction accuracy. Furthermore, coherency constraints \cite{taieb2017coherent} also add more complication to the prediction model.
The straight-forward methods to utilize hierarchical structure include \textit{bottom-up} forecasting and \textit{top-down} forecasting. As their names suggest, these methods make individual predictions at the bottom or upper levels and then aggregate according to the hierarchical structure. Although these methods naturally satisfy coherency, they are unable to consider statistics of all levels for prediction at the same time.  Specifically, when forecasting from a single level of the aggregation structure, these methods either aggregate or disaggregate to obtain forecasts of all the rest levels.  In fact, the statistical characteristics at different levels can be drastically distinct, e.g., time series of upper levels tend to be more stationary, while those at the bottom levels are often more fluctuant, for which flexible adaptations at different levels is desirable to better utilize the information at all levels.

To address these problems, the reconciliation method, revising the predictions for coherency after forecasting all of the series individually, has become popular in the recent research \cite{hyndman2011optimal}.  These works usually follow the two-stage approach: First, independently forecast all the series (i.e., generate incoherent base forecasts); Second, reconcile the base forecast by applying forecast combination using a bottom-up procedure. In other words, the base forecasts are adjusted so that they become coherent. MinT \cite{wickramasuriya2019optimal} is one of the representative methods, which reconciles the base forecasts via the optimal combination with minimum variance among all unbiased revised forecasts. 

Most previous works that reconcile forecasts of all levels to ensure coherency face the following challenges: \textbf{(i)} The base forecast is obtained independently without any shared information from other time series.  \textbf{(ii)} State-of-the-art methods rely on strong statistical assumptions, such as, unbiased forecasts and Gaussian noises, but the data distributions in real-world are mostly non-Gaussian/non-linear across the hierarchy, which calls for a method that can project data distribution into Gaussian space where tractable methods can be applied. \textbf{(iii)} The two-stage approaches reconcile the base forecasts without any regard to the learned model parameters, and thus cannot fully utilize the power of deep parametric models, resulting in the lack of information sharing between the process of prediction and reconciliation. \textbf{(iv)} Most methods only focus on generating point estimates, but the probabilistic forecasts are often necessary in practice to facilitate the subsequent decision-making processes.

In this paper, we present a novel end-to-end approach that tackles forecasting and reconciliation simultaneously for hierarchical time series, without requiring any explicit post-processing step, by combining the recently popular \textit{autoregressive transformer} \cite{katharopoulos2020transformers} and \textit{conditioned normalizing flow} \cite{korshunova2018conditional} to generate the coherent probabilistic forecasts with the state-of-the-art performance.

Specifically, we \textbf{first} obtain the base forecast via the autoregressive transformer, modeling the multivariate time series of all-levels. Using encoder-decoder transformer structure, which has been successful in recent advances in the multivariate time series forecasting \cite{zhou2021informer}, we achieve the information fusion of all levels in hierarchy via globally shared parameters, while benefiting from the representation power of the autoregressive transformer model.  Transformer models have also shown superior performance in capturing long-range dependency over RNN-based models.   \textbf{Second}, we reconcile the base forecasts into coherent forecasts via conditioned normalizing flow (CNF) \cite{korshunova2018conditional} with bottom-up aggregation matrix. 
Since we need to model complex statistical properties in hierarchical data for our probabilistic forecasting, normalizing flow (NF), a proven powerful density approximator, becomes our natural choice. 
Furthermore, by extending NF to the conditional form, we can incorporate base forecasts from all levels into CNF as additional conditions for the latent space, leveraging the information available across all levels for reconciliation, while modeling the non-Gaussian distributions and non-linear correlations in the hierarchy to obtain probabilistic forecasts. \textbf{Finally},  throughout the whole process, we combined forecasting and reconciliation simultaneously for end-to-end training, while ensuring the continuity and differentiability at all steps.
Moreover, our framework is able to accommodate different loss functions besides log-likelihood, and by sampling from the forecast distribution, sufficient statistics can also be obtained via the empirical distribution to facilitate even more complicated optimization objectives.

\subsubsection*{\textbf{Our Contributions}}
We summarize our contributions as follows:
\begin{itemize}
\item A multivariate autoregressive transformer architecture, modeling each time series simultaneously via globally shared parameters
\item A novel reconciliation method via CNF with bottom-up aggregation matrix to integrate information of all level in the hierarchy for coherent probabilistic forecast without relying on any statistical assumption
\item An end-to-end learning framework of hierarchical time series achieving forecasting and reconciliation simultaneously, without requiring any explicit post-processing step
\item Extensive experiments on real-world hierarchical datasets from various industrial domains and real-life deployment for traffic forecasting on Alipay's application servers
\end{itemize}

In the following content, we introduce the involved background and review the related work in Section 2 and 3, respectively. We then present our proposed method and describe the procedure of training and inference in Section 4. Finally in Section 5, we analyse non-Gaussian/non-linear properties of real-world hierarchical data and demonstrate the advantages of our approach through experiments and ablation study. 

\section{Background}
\subsection{Hierarchical Time Series and Reconciliation}
The hierarchical time series can be expressed as a tree structure  (see Figure 1) with linear aggregation constraints, represented by aggregation matrix $S\in \mathbb{R}^{n \times m}$ ($m$ is number of bottom-level nodes, and $n$ is total number  of nodes).  Each node in the hierarchy represents one time series, to be predicted over a time horizon. 
\begin{figure}[h]
  \centering
  \includegraphics[width=\linewidth]{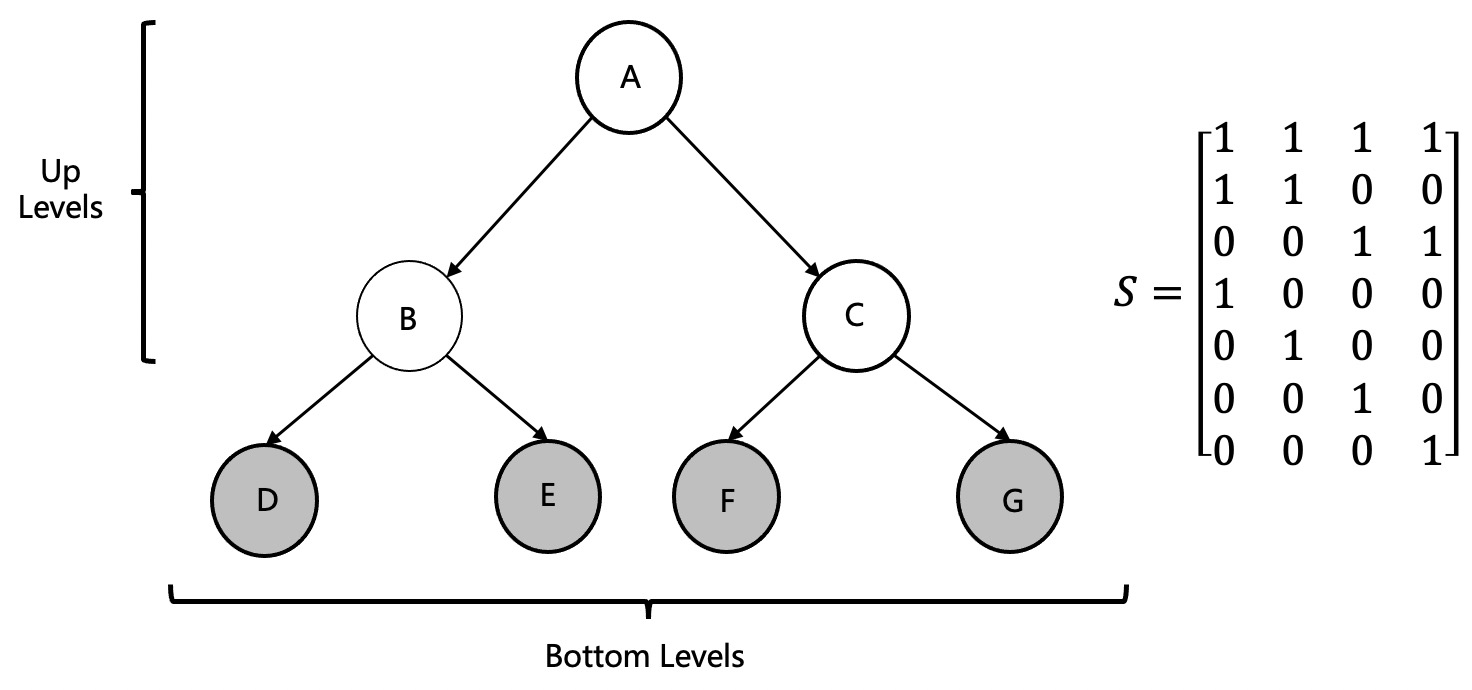}
  \caption{Example of hierarchical time series structure for n=7 time series with m=4 bottom-level series and r=3 upper-level series}
\end{figure}
Given a time horizon $t\in \{1,2,...,T\}$, we use $y_{t,i}\in \mathbb{R}$ to denote the values of a multivariate hierarchical time series, where $i\in \{1,2,...,n\}$ is the index of the individual univariate time series.  Here we assume that the index $i$ of the individual time series abides by the level-order traversal of the hierarchical tree, going from left to right at each level, and we use $x_{t,i}$ to denote time-varying covariate vectors associated with each univariate time series $i$ at time step $t$. 

In our tree hierarchy, the time series of leaf nodes are called the \textit{bottom-level} series $b_{t}\in \mathbb{R}^{m}$, and those of the remaining nodes are termed the \textit{upper-level} series $u_{t}\in \mathbb{R}^{r}$. Obviously, the total number of nodes $n=r+m$, and $y_{t} =[u_{t},b_{t}]^{T} \in  \mathbb{R}^{n} $ contains observations at time $t$ for all levels , which satisfy (by an aggregation matrix $S\in \{0,1\}^{n \times m}$):   
\begin{equation}
y_{t} =[u_{t},b_{t}]^{T} \quad \Leftrightarrow \quad y_{t}=Sb_{t} \quad	\Leftrightarrow \quad \begin{bmatrix} u_{t} \\ b_{t}\end{bmatrix} =Sb_{t}  
\mbox{ ,}\end{equation}
for every time step $t$. For example, in Figure 1, the total number of series in the hierarchy is $n=r+m=3+4 $, i.e., the number of series at the bottom-level is $m=4$ and the number of upper-level series is $r=3 $. For every time step $t$, $u_{t}=[y_{A,t},y_{B,t},y_{C,t}] \in \mathbb{R}^{3} $ and $b_{t}=[y_{D,t},y_{E,t},y_{F,t},y_{G,t}] \in \mathbb{R}^{4}$, the $y_{A,t}=y_{B,t}+y_{C,t}=y_{D,t}+y_{E,t}+y_{F,t}+y_{G,t} $, and the aggregation matrix $S\in \{0,1\}^{7 \times 4}$.

\subsubsection*{\textbf{Reconciliation of Hierarchical Forecasting.}} 
For given multivariate hierarchical time series, we first consider ignoring the aggregation constraints and forecasting all time series separately, which are called \textit{base forecasts} denoted by $\hat{y}_{t} \in \mathbb{R}^{n}$, where $h$ is the forecast horizon.
Then all forecasting approaches for hierarchical structures can be represented as 
\begin{equation}
\tilde{y_{t}}=\bm{SP}\hat{y_{t}}
\mbox{ ,}\end{equation}
where $\bm{P} \in \mathbb{R}^{m \times n}$ is a matrix that projects the base forecasts (of dimension $n$) into the bottom-forecast(of dimension $m$), which is then summed up by the aggregation matrix  $\bm{S}\in \{0,1\}^{n \times m}$ using the aggregation structure to produce a set of \textit{coherent forecasts} $\tilde{y}_{t} \in \mathbb{R}^{n}$, which satisfy the aggregation constraints. Equation 2 can be easily extended to $h$-period-ahead forecast
\begin{equation*}
\tilde{y}_{t+h}=\bm{SP}\hat{y_{t+h}}
\mbox{ .}
\end{equation*}
Typically, the base and coherent forecasts can be linked by Equation 2 above, which is also called \textbf{\textit{reconciliation process}}. A main effort of the state-of-the-art methods is to improve the reconciliation process, and in Section 3, we will review the related work in this domain.

\subsection{Autoregressive Transformer}
Recently, the encoder-decoder transformer structure has been highly successful in advancing the research on multivariate time series forecasting, enabled by its multi-head self-attention mechanism to capture both long- and short-term dependencies in time series data \cite{zhou2021informer,li2019enhancing}. Meanwhile, combined with classical autoregressive models for time series, we can extend the transformer structure to an autoregressive deep learning model using causal masking, which preserves the autoregressive property by utilizing a mask that reflects the causal direction of the progressing time, i.e., masking out data from future \cite{katharopoulos2020transformers}.

Specifically, given $\mathcal D$, defined as a batch of time series $\bm{Y}=[\bm{y}_{1},...,{\bm{y}_{T}}] \in \mathbb{R}^{T \times D}$, the transformer takes in the above sequence $\bm{Y}$, and then transforms this into dimension $l$ (i.e., $dl$) to obtain the query matrix $\bm{Q}$, the key matrix $\bm{K}$ and the value matrix $\bm{V}$ as follows:
\begin{equation}
    \begin{aligned}
      \bm{Q}=\bm{Y} \bm{W}_{l}^{Q} \quad
      \bm{K}=\bm{Y} \bm{W}_{l}^{K} \quad
      \bm{V}=\bm{Y} \bm{W}_{l}^{V} \mbox{ ,}
    \end{aligned}
\end{equation}
where $\bm{Q} \in \mathbb{R}^{dl \times dh}$, $\bm{K} \in \mathbb{R}^{dl \times dh}$, $\bm{V} \in \mathbb{R}^{dl \times dh}$ (we use $dh$ to denote the length of time steps), and $\bm{W}_{l}^{Q}$, $\bm{W}_{l}^{K}$, $\bm{V}_{l}^{V}$ are learnable parameters. 
After these linear transformation, the scaled dot-product attention computes the sequence of vector outputs via:
\begin{equation}
    \bm{S}=\text{Attention}(\bm{Q},\ \bm{K},\ \bm{V})=\operatorname{softmax}\left(\frac{\bm{Q} \bm{K}^{T}}{\sqrt{d_{K}}} \cdot \bm{M}\right ) \bm{V}
 \mbox{ ,}\end{equation}
where the mask matrix $M$ can be applied to filter out right-ward attention (or future information leakage) by setting its upper-triangular elements to $-\infty$ and normalization factor $d_{K}$ is the dimension of the $W_{h}^{K}$ matrix.  Finally, all outputs $S$ are concatenated and linearly projected again into the next layer.

Transformer is commonly used in an encoder-decoder architecture, where some warm-up time series as context are passed through the encoder to train the decoder and autoregressively obtain predictions. 

In this work, we employ the multivariate autoregressive transformer architecture, to model each time series simultaneously via globally shared parameters. By doing so, we achieve the information fusion of all levels in hierarchy to generate the base forecasts. In fact, in our case, \textit{base forecasts} from the base model do not directly correspond to un-reconciled forecasts for the base series, but rather represent predictions of an unobserved latent states, which are used for subsequent density estimations.

\subsection{Density Estimation Via Normalizing Flow} 
Normalizing Flows (NF) \cite{papamakarios2019normalizing}, which learn a distribution by transforming the data to samples from a tractable distribution where both sampling and density estimation can be efficient and exact, have been proven to be powerful density approximators.
The change of variables formula (Equation 5 below) empowers the computation of exact likelihood \cite{kobyzev2020normalizing}, which is in contrast to other powerful density estimator such as Variational Autoencoders or Generative Adversarial Networks.  Impressive estimation results, especially in the field of nonlinear high-dimensional data generation, have lead to great popularity of flow-based deep models \cite{kingma2018glow}. 

NF are invertible neural networks that typically transform isotropic Gaussians to characterize a more complex data distribution, projecting from  $\mathbb{R}^{D}$ to $\mathbb{R}^{D}$ such that densities $p_{Y}$ on the input space $Y  \in \mathbb{R}^{D}$ are transformed into some tractable distribution $p_{Z}$ (e.g., an isotropic Gaussian) on space $Z  \in \mathbb{R}^{D}$.  This mapping function, $f:Y \rightarrow Z$ and inverse mapping function, $f^{-1}:Z \rightarrow Y$ are composed of a sequence of bijections or invertible functions, and we can express the target distribution densities $p_{Y}(\bm{y})$ by
\begin{equation}
p_{Y}(\bm{y})=p_{Z}(\bm{z})|det(\frac{\partial	f(\bm{y})}{\partial \bm{y}})| \mbox{ ,}\end{equation}
where $\partial f(y)/\partial y$ is the Jacobian of $f$ at $y$.  NF have the property that the inverse $y=f^{-1}(\bm{z})$ is easy to evaluate and computing the Jacobian determinant takes $O(D)$ time.

For mapping functions $f$, the bijection introduced by RealNVP architecture (the coupling layer) \cite{dinh2016density} satisfies the above properties, leaving part of its inputs unchanged, while transforming the other part via functions of the un-transformed variables (with superscript denoting the coordinate indices)
\begin{equation}
  \left\{ 
    \begin{array}{lr}   
        y^{1:d}=z^{1:d} \\
        y^{d+1:D}=z^{d+1:D} \odot exp(s(z)^{1:d}+t(z^{1:d})) 
    \end{array}
 \right.  
\mbox{ ,}\end{equation}
where $\odot$ is an element wise product, $s()$ is a scaling and $t()$ is a translation function from $\mathbb{R}^{D} \mapsto \mathbb{R}^{D-d}$, using neural networks. To model a nonlinear density map $f(x)$, a number of coupling layers, mapping $\mathcal{Y} \mapsto \mathcal{Y_{1}} \mapsto ...\mapsto \mathcal{Y_{K-1}}  \mapsto \mathcal{Y_{K}} \mapsto \mathcal{Z}$, are composed together with unchanged dimensions.

\subsubsection*{\textbf{Conditional Normalizing Flow.}}
Inspired by the conditional extension of NF \cite{korshunova2018conditional}, with the conditional distribution $p_{Y}(\bm{y}|\bm{h})$, we realize that the scaling and translation function approximators do not need to be invertible \cite{korshunova2018conditional}, which means we can make the transformation dependant on condition $\bm{h} \in \mathbb{R}^{H}$.  Implementing $p_{Y}(\bm{y}|\bm{h})$ on $\bm{h}$ is straight-forward: we concatenate $h$ to both the inputs of the scaling and translation function approximators of the coupling layers, i.e., $s(concat(\bm{z}^{1:d},\bm{h}))$ and $t(concat(\bm{z}^{1:d},\bm{h}))$, which are modified to map $\mathbb{R}^{d+H} \mapsto \mathbb{R}^{D-d}$. 

In our work, we incorporate base forecasts (Section 2.2), from the outputs of Autoregressive Transformer at all levels into CNF as additional condition in the latent space. In Section 4, we will detail this process.

\section{Related work}
Existing hierarchical time series forecasting methods mainly follow the two-stage approach: \textbf{(i)} Obtain each $h$-period-ahead base forecasts $\hat{\bm{y}}_{T+h}$ independently; \textbf{(ii)} Reconcile the base forecasts by the reconciliation process (Equation 2) to obtain the coherent forecasts $\tilde{\bm{y}}_{T+h}$.  

This approach has the following advantages:\textbf{(1)} The forecasts are coherent by construction; \textbf{(2)} The combination of forecasts from all levels is applied via the projection matrix $\bm{P}$, where information from all levels of hierarchy is incorporated simultaneously.  The main work of the current state-of-the-art method is to improve the reconciliation process, which will be reviewed in this section.

Wickramasuriya \textit{et al.} proposed \textbf{Mint} \cite{wickramasuriya2019optimal} to optimally combine the base forecasts.  Specifically, assuming the base forecasts $\hat{y}_{T+h}$ are unbiased, Mint computes the projection matrix $\bm{P}=(S^{T}W^{-1}_{h}S)^{-1} (S^TW^{-1}_{h})$ giving the minimum variance unbiased revised forecasts, i.e., minimizing $tr[SPW_hP^TS^T]$ with constraint $SPS=S$, where  $W_{h}$ is the covariance matrix of the $h$-period-ahead forecast errors $\hat{\varepsilon}_{T+h}=y_{T+h}-\hat{y}_{T+h}$ ($\bm{S}$ and $\bm{P}$ are matrices defined in eq. (2)).  However, the covariance of errors $W_{h}$ is hard to obtain for general $h$ and the strong assumption of unbiased base forecasts is normally unrealistic.

The unbiased assumption is relaxed in \textbf{Regularized Regression for Hierarchical Forecasting} \cite{ben2019regularized}, which also follows the two-stage approach and seeks the revised forecasts with the trade-off between bias and variance of the forecast by solving an empirical risk minimization (\textbf{ERM}) problem.

\textbf{Probabilistic Method for Hierarchical Forecasting} \cite{panagiotelis2020probabilistic} also employs the two-stage approach, but in contrast to the above methods, it considers forecasting probability distributions rather than just the means (point forecasts).  This probabilistic method starts by generating independent forecasts of the conditional marginal distributions, followed by samplings from the above distributions as the base forecasts, which are then reconciled using Equation 2.  In this approach, existing reconciliation methods for point forecasts can be extended to a probabilistic setting.  But the reconciliation is only applied to the samples rather than the forecast distribution, which creates another level and uncertainty with unsalable computation complexity.

One typical work in end-to-end modeling of hierarchical time series is \textbf{Hier-E2E} \cite{rangapuram2021end}, where base forecasts are obtained using DeepVAR \cite{salinas2019high} with diagonal Gaussian distribution, followed by reconciliation using a closed-form formulation of an optimization problem, i.e., minimizing the errors of base forecast $\hat{y}$ and reconciling forecast $\tilde{y}$ subject to coherent constraints of hierarchical structure. The reconciliation process is as follows:
\begin{gather*}
\tilde{y_{t}}=M\hat{y_{t}} \\
M:=I-A^{T}(AA^T)^{-1}A\mbox{ ,}
\end{gather*}
where $M$ is a fixed matrix and $A$ is a structure matrix from the upper half of the aggregated matrix.  In contrast to the previous work, the model no longer has a relationship with the predicted value $y$. The matrix $M$ is time-invariant and can be computed offline, prior to the training. This reconciliation approach is essentially a fine-tuning of the base forecasts under the coherence constraint, and the learned model parameters are not used to revise the base forecast in the reconciliation stage.

Considering the pros and cons of the above models, we combine the advantages and propose a novel end-to-end model that not only obtains the coherent probabilistic forecasts, but also achieves the reconciliation in the forecast distributions (rather than just on samples), ensuring that the reconciliation is related to the predicted value $y$ via integrating information from all levels, to improve the overall performance.

\section{Methodology}
In this section, we explain our approach that combines the multivariate autoregressive transformer and CNF for coherent probabilistic forecasting.

\subsection{Model}
A schematic overview of our hierarchical end-to-end architecture can be found in Figure 2.
\subsubsection{\textbf{Multivariate Autoregressive Transformer} \label{sec:model_1}} 
\begin{figure*}[h]
  \centering
  \includegraphics[width=\linewidth]{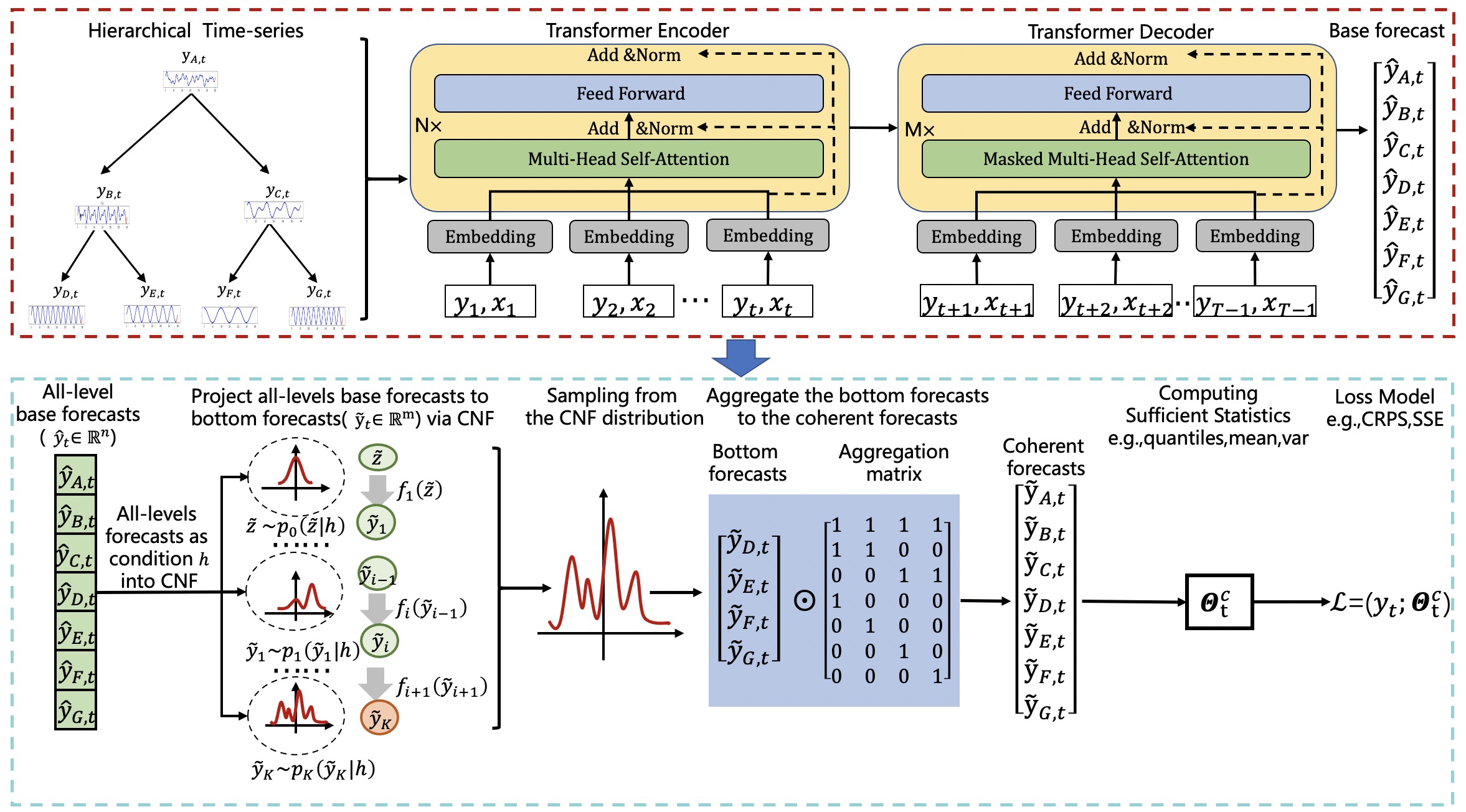}
  \caption{Model Architecture. The red dashed line represents the multivariate autoregressive transformer module ($\S$ \ref{sec:model_1}) and the blue dashed line highlights the reconciliation method via CNF with bottom-up aggregation matrix ($\S$ \ref{sec:model_2}). We can compute sufficient statistics from the samples via the empirical distribution to facilitate more complicated optimization objectives.}
\end{figure*}

The encoder -decoder transformer architecture has been highly successful in advancing the research on multivariate time series forecasting, enabled by its multi-head self-attention mechanism to capture both long- and short-term dependencies, and can be further extended to have autoregressive properties by using \textit{causal masking}.  These advantages motivate the use of \textit{multivariate autoregressive transformer} as our building block to better capture patterns (e.g., trends and cycles) inside each individual time series.  Note that the global parameters of the transformer are shared across different time series to exploit common patterns over the entire history.

We denote the entities of a hierarchical time series by $y_{t,i}\in \mathbb{R}$ for $i\in \{1,2,...,n\}$, where $t$ is the time index.  We consider
time series with $t\in [1,T]$, sampled from the complete history of our data, where for
training we split the sequence by some context window $[1,t_{0})$ and prediction window $[t_{0},T]$.  We use $x_{t,i}$ to denote time-varying covariate
vectors associated with each univariate time series ${i}$ at time step $t$.

In the encoder-decoder transformer architecture, the encoder embeds $\bm{y}_{1:t_{0}-1}$ and the decoder outputs the base forecast of all levels as condition for the density estimations over $\bm{y}_{t_{0}:T}$ via a masked attention module:
\begin{equation}
    \begin{aligned}
        \bm{h}_{t_{0}}=\textsl{TransformerEncoder}(concat(\bm{y}_{1:t_{0}-1},\bm{x}_{1:t_{0}-1});\theta) \\
        \bm{\hat{y}}_{t}=\textsl{TransformerDecoder}(concat(\bm{y}_{t-1},\bm{x}_{t-1}),\bm{h}_{t_{0}}; \phi)
        \mbox{ ,}
    \end{aligned}
\end{equation}
where $\theta$ and $\phi$ are parameters of the transformer's encoder and decoder, respectively.  These parameters are shared globally, achieving information fusion across all levels in hierarchy to generate the \textit{base forecasts} $\bm{\hat{y}}_{t}$.

During training, care has to be taken to prevent using information from future.  Specifically, to ensure the autoregressive property of the model, we employ a mask that reflects the causal direction of the progressing
time, i.e. to mask out future time points. 

Note that, in our case, base forecasts $\bm{\hat{y}}_{t}$ do not directly correspond to un-reconciled forecasts for the base series, but rather represent predictions of an unobserved latent states, which can be used for subsequent density estimations.
The transformer allows the model to access any part of the historic time series regardless of temporal distance and thus is potentially able to generate better condition $\bm{h}_{t} \in \mathbb{R}^{H}$ for the subsequent density estimations.

\subsubsection{\textbf{Reconciliation via Conditional Normalizing Flow} \label{sec:model_2}}
\ 
Next we describe how to obtain the \textit{coherent probabilistic forecasts}, given the base forecasts described above from the autoregressive transformer.

To estimate the probability density of data (in order to obtain probabilistic forecast), one straight-forward method is to use parameterized Gaussian distribution, but as mentioned above, the real-world hierarchical data are mostly non-Gaussian/non-linear. Equipped with the powerful density approximator, NF, we are able to tackle this challenge, capturing the nonlinear relationships among all levels in hierarchy. 

According to the definition of Equation 2, the reconciliation takes a projection matrix $\bm{P} \in \mathbb{R}^{m \times n}$, projecting from the base forecasts (of dimension $n$) into the bottom-forecasts (of dimension $m$).  In our approach, this projection is replaced using \textit{conditional normalizing flow} (CNF), i.e., the conditional joint distribution $p_{Y}(\tilde {\bm{y}}_{t}|\bm{h}_{t})$, where $\bm{h}_{t}$ is the condition (base forecasts $\hat{\bm{y}_{t}}$ in our case) and the current $\bm{\tilde{y}}_{t}$ is the reconciled bottom-forecasts (of dimension $m$).

In the Real-NVP architecture,  we extend Equation 6 by concatenating condition $\bm{h}_{t}$ to both the inputs of the scaling and translation function approximators of the coupling layers as follows:
\begin{equation}
  \left\{ 
    \begin{array}{lr}   
        \bm{y}^{1:d}=\bm{z}^{1:d} \\
        \bm{y}^{d+1:D}=\bm{z}^{d+1:D} \odot exp(s(\bm{z}^{1:d},\bm{h})+t(\bm{z}^{1:d},\bm{h})) 
    \end{array}
 \right.  
\mbox{ ,}\end{equation}
where $\bm{z}$ is a noise vector sampled from an isotropic Gaussian, functions $s$ (scale) and $t$ (translation) are usually deep neural networks, which as mentioned above, need not to be invertible.

To obtain an expressive distribution representation, we can stack \textit{K} layers of conditional flow modules (Real-NVP), generating the conditional distribution of the future sequences of all time series in hierarchy, given the past time $t \in [1,t_{0})$ and the covariates in $t \in [1,T]$.  Specifically, it can be written as a product of factors (as an autoregressive model):
\begin{equation}
p_{Y}(\bm{\tilde{y}}_{t_{0}:T}|\bm{y}_{1:t_{0}-1},\bm{x}_{1:T};\theta,\phi,\psi)=\prod_{t=t_{0}}^{T} p_{Y}(\bm{\tilde{y}}_{t}|\bm{h}_{t};\theta,\phi,\psi)
\mbox{ ,}\end{equation}
where $\theta$ and $\phi$ are parameters of the transformer and $\psi$ is the parameter of CNF.

Then we can generate directly a set of Monte Carlo samples from the the above conditional joint distribution $p_{Y}(\bm{\tilde{y}}_{t}|\bm{h}_{t})$ as the reconciled bottom forecasts, e.g., in Figure 2, $[\tilde{y}_{D,t},\tilde{y}_{E,t},\tilde{y}_{F,t},\tilde{y}_{G,t}]$. According to Equation 2 ($\tilde{y_{t}}=\bm{SP}\hat{y_{t}}$), the bottom forecasts are multiplied by the aggregation matrix $S$ to obtain the coherent probabilistic forecasts $\tilde{\bm{y}}_{t}$ of all levels.

Our approach combines prediction and reconciliation into a unified process using deep parametric models, facilitating the sharing of information in process described above via global parameters.  Our reconciliation method not only ensures hierarchical coherence constraints, but also dynamically revises the base forecast via integrating information of all levels, to improve the overall performance.  Unlike current probabilistic method for hierarchical forecasting, we apply the projection matrix $P$ on the conditional joint distribution of CNF, rather than samples from the distribution, resulting in exact estimation with higher efficiency.

Moreover, our framework can accommodate different loss functions besides log-likelihood, and by sampling from the forecast distribution, we can also obtain sufficient statistics via the empirical distribution to facilitate more complicated optimization objectives.

\subsection{Training}
During training, our loss function is directly computed on the coherent forecast samples.
Specifically, given $\mathcal D$, defined as a batch of time series $Y:=\{y_{1},y_{2},...,y_{T}\}$, and the associated covariates $X:=\{x_{1},x_{2},...,x_{T}\}$, we can maximize the likelihood given by Equation 9 via SGD using Adam, i.e.
\begin{equation}
\begin{aligned}
      \mathcal{L}&=
      \frac{1}{|\mathcal D|T}\prod_{x1:T,y1:T \in \mathcal D}\prod_{t=1}^{T} p_{Y}(y_{t}|y_{1:t-1};x_{1:t},\theta,\phi,\psi) \\
      &=\frac{1}{|\mathcal D|T}\prod_{x1:T,y1:T \in \mathcal D}\prod_{t=1}^{T} p_{Y}(y_{t}|h_{t},\theta,\phi,\psi)
\end{aligned}
 \mbox{ ,}\end{equation}
where the globally shared parameters ($\theta$, $\phi$) and $\psi$ are from the transformer
module and the CNF module, respectively.

Please note that we can easily obtain $\theta$, $\phi$, and $\psi$ for other loss functions such as quantile loss, CRPS (continuous ranked probability score) or any other metrics preferred in the forecasting community, as long as we can compute the sufficient statistics from the Monte Carlo samples $\{\tilde{y_{t}}\}$ via the empirical distribution function.
\vspace*{-\abovedisplayskip}
\subsection{Inference}
During inferencing, we can predict using the autoregressive transformer in a step-by-step fashion over the time horizon. Specifically, we first generate the base forecasts $\hat{y}_{t}$ using autoregressive transformer for one time step using the covariate vector $x_{1:t-1}$ and the observed value $y_{1:t-1}$. Then we can incorporate base forecasts $\hat{\bm{y}}_{t}$ from all levels into CNF as additional condition $\bm{h}_{t}$ in the latent space to model conditional joint distribution $p_{Y}(\tilde {\bm{y}}_{t}|\bm{h}_{t})$.  Lastly, we directly obtain a set of Monte Carlo samples from the distribution $p_{Y}(\tilde {\bm{y}}_{t}|\bm{h}_{t})$ as the reconciled bottom forecasts, which are multiplied by the aggregation matrix $S$ to obtain the coherent probabilistic forecasts $\tilde{\bm{y}}_{t}$.
Note that we can repeat the above procedure for the $h$-period-ahead forecasting to obtain a set of coherent forecasts $\{\tilde{y_{T}},\tilde{y_{T+1}},...,\tilde{y_{T+h}}\}$.

\section{Experiments \label{sec:experiment}}
In this section, we conduct extensive empirical evaluations on four real-world hierarchical datasets from different industrial domains, including three public datasets and one dataset collected from the application servers of Alipay, which is in the internal review process for releasing publicly.

\subsection{Datasets}
\subsubsection*{\textbf{Datasets}}
The three public datasets include Tourism \cite{bushell2001tourism}, Tourism-L \cite{wickramasuriya2019optimal}, and Traffic \cite{cuturi2011fast}.  The new dataset is the data traffic collected from Alipay's application servers, where our method is deployed to forecast data traffic. 
Three real-world hierarchical datasets and a dataset of web traffic from application servers from Alipay are listed below for our experiments:
\begin{itemize}
\item{
Traffic \cite{cuturi2011fast} provides the occupancy rate (between 0 and 1) of 963 car lanes of San Francisco bay area freeways. We aggregate sub-hourly data to obtain daily observations for one year and generate a 207-series hierarchy using the same aggregation strategy as in \cite{ben2019regularized}, which is divided into 4 levels.  Bottom-level contains 200 series, aggregated-levels contain 7 series in total, and the prediction length is 1. }
\item{
Tourism \cite{bushell2001tourism,athanasopoulos2009hierarchical} includes an 89-series geographical hierarchy with quarterly observations of Australian tourism flows from 1998 to 2006, which is divided into 4 levels.  Bottom-level contains 56 series, aggregated-levels contain 33 series, and the prediction length is 8.  This dataset is frequently referenced in hierarchical forecasting studies \cite{hyndman2018forecasting,taieb2021hierarchical}. }
\item{
Tourism-L \cite{wickramasuriya2019optimal} is a larger, more detailed version of Tourism, which contains 555 total series in a grouped structure and 228 observations; This dataset has two hierarchies, i.e., based on geography and based on purpose-of-travel, respectively, sharing a common root, which is divided into 4 or 5 levels.  Bottom-level contains 76 or 304 series, aggregated-level contains 175 series, and the prediction length is 12. }
\item{
Server-traffic is the data traffic collected
from Alipay’s application servers over the past 90 days, where our method is deployed for data traffic forecasting. This dataset has the three levels: Bottom-level contains 35 series, aggregated-levels contain 4 series and the prediction length is 8.}
\end{itemize}




\subsection{SOTA Methods and Evaluation Metrics}

\begin{table*}[h]
  \label{table:experiment}
  \setlength\tabcolsep{11pt} 
  \begin{spacing}{1}
  \begin{center}
        \begin{tabular}{cccccl}
    \toprule
    \textbf{Method} & Traffic & Tourism & Tourism-L & Server-Traffic\\
    \midrule
   ARIMA-NaiveBU & 0.0808 & 0.1138 & 0.1741 &  0.3834 \\
   ETS-NaiveBU & 0.0665 & 0.1008 & 0.1690 &  0.3899 \\
   ARIMA-MinT-shr & 0.0770 & 0.1171 & 0.1609 &  0.2713 \\
   ARIMA-MinT-ols & 0.1116 & 0.1195 & 0.1729 &  0.2588 \\
   ETS-MinT-shr & 0.0963 & 0.1013 & 0.1627 &  0.3472 \\
   ETS-MinT-ols & 0.1110 & 0.1002 & 0.1668 &  0.2652 \\
   ARIMA-ERM & 0.0466 & 0.5887 & 0.5635 &  0.2320 \\
   ETS-ERM & 0.1027 & 2.3755 & 0.5080 &  0.2501 \\
   DeepVAR-lowrank-Copula & 0.0583±0.0071 & 0.0991±0.0083 & 0.1781±0.0093 &  0.1125±0.0041 \\
   Hier-E2E & 0.0376±0.0060 & 0.0834±0.0052 & 0.1520±0.0032 &  0.0530±0.0127 \\
   Hier-Transformer-CNF(\textbf{Ours}) & \textbf{0.0217±0.0055} & \textbf{0.0611±0.0077} & \textbf{0.1135±0.0059} & \textbf{0.0314±0.0067} \\
  \bottomrule
  Multivariate Autoregressive Transformer  & 0.0377 & 0.0815 & 0.1471 &  0.0417 \\
  \bottomrule 
  \end{tabular}
  \end{center}
  \end{spacing}
   TABLE \uppercase\expandafter{\romannumeral1}: CRPS values(lower is better) averaged over 5 runs. We conduct the ablation study in Multivariate Autoregressive Transformer. State-of-the-art methods except for DeepVAR-lowrank-Copula and Hier-E2E produce consistent results over multiple runs.
\end{table*}

\subsubsection*{\textbf{State-of-the-art Methods}}
We conduct the performance comparison against state-of-the-art reconciliation algorithms including MinT \cite{wickramasuriya2019optimal}, ERM \cite{ben2019regularized}, and Hier-E2E \cite{rangapuram2021end}, along with other classical baselines, including bottom-up (NaiveBU) approach that generates univariate point forecasts for the bottom-level time series independently followed by the aggregation according to the coherent constraints to obtain point forecasts for the aggregated series. 
The specific details are as follows:
\begin{itemize}
\item{
\textbf{ARIMA-NaiveBU} and \textbf{ETS-NaiveBU}: \textit{NaiveBU} generates univariate point forecasts for the bottom-level time series independently and then sums them according to the hierarchical constraint to obtain point forecasts for the aggregate series. Specifically, we use ARIMA and ETS with auto-tuning for base forecast in \textit{NaiveBU} with the R package \textit{hts}.}
\item{
\textbf{ARIMA-MinT-shr}, \textbf{ARIMA-MinT-ol}, \textbf{ETS-MinT-shr}, \textbf{ETS-MinT-ols} :  For \textit{MinT}, we use the covariance matrix with shrinkage operator (\textbf{MinT-shr}) and the diagonal covariance matrix corresponding to ordinary least squares weights (\textbf{MinT-ols}) \cite{wickramasuriya2019optimal}.  In addition, we also use ARIMA and ETS for auto-tuning of base forecasts in \textit{Mint} with the R package \textit{hts}.}
\item{
\textbf{ARIMA-ERM} and \textbf{ETS-ERM}: \textit{ERM} is from Regularized Regression for Hierarchical Forecasting \cite{ben2019regularized} in related work discussed in the previous section and we also use ARIMA and ETS for auto-tuning of base forecasts in \textit{ERM} with the R package \textit{hts}.}
\item{
\textbf{DeepVAR-lowrank-Copula}: \textit{DeepVAR} is a multivariate, nonlinear generalization of classical autoregressive model \cite{salinas2019high}.  We use the vanilla model with no reconciliation, parameterizing a low-rank plus diagonal convariance via Copula process.  We set the rank of low-rank convariance to be 5. The code is from the Gluonts implementatin on https://github.com/awslabs/gluon-ts.}
\item{
\textbf{Hier-E2E}: In \textit{Hier-E2E} \cite{rangapuram2021end}, we use the DeepVAR\cite{salinas2019high} with the diagonal covariance matrix to obtain the base forecast, and use
pre-calculated reconciliation matrix $M$ to project into the coherent
subspace. The code is from the authors’ implementation on https://github.com/rshyamsundar/gluonts-hierarchical-ICML-2021.}
\end{itemize}

\subsubsection*{\textbf{Evalution Metrics}}
In the stage of evaluation, we use the \textit{Continuous Ranked Probability Score} (CRPS), which measures the compatibility of a cumulative distribution function $\hat{F}^{-1}_{t,i}$ for time series $i$ against the ground-truth observation $y_{t,i}$, to estimate the accuracy of our forecast distributions.  CRPS can be defined as 
\begin{equation}
CRPS(\hat{F}_{t},y_{y}):=\sum_{i} \int_{0}^{1}QS_{q}(\hat{F}^{-1}_{t,i}(q),y_{t,i}) \ dq
 \mbox{ ,}\end{equation}
where $QS_{q}$ is the quantile score for the $q$-th quantiles:
\begin{equation*}
QS_{q}=2(\mathds{1}\{y_{t,i} \leq \hat{F}^{-1}_{t,i}(q)\}-q)(\hat{F}^{-1}_{t,i}(q)-y)
\end{equation*}

We use the discrete version in our experiment, with the integral in Equation 11 replaced by the weighted sum over the quantile set, and we use the quantiles ranging from 0.05 to 0.95 in steps of 0.05.

\subsection{Experiment Environment}
All experiments run on the Linux server(Ubuntu 16.04) with the Intel(R) Xeon(R) Silver 4214 2.20GHz CPU, 16GB memory, and the signle Nvidia V-100 GPU.

\subsection{Experiment Results}
For evaluation, we generate 200 samples from our probabilistic model to create an empirical predictive distribution. We run our method 5 times and report the mean and standard deviation of CRPS scores.  Table 1 shows the performances of different approaches on the four hierarchical datasets, and we can observe that our proposed approach (Hier-Transformer-CNF) achieves all best results, with significant improvements of accuracy on most datasets.

\subsection{Ablation Study}
\label{sec:albation study}
To further demonstrate the effectiveness of the designs described in Section 4, we conduct ablation studies using the variant of our model and the related models. The results of each aggregation level are shown in Figure 7.
\begin{figure}[h]
  \centering
  \includegraphics[width=\linewidth,height=0.6\linewidth]{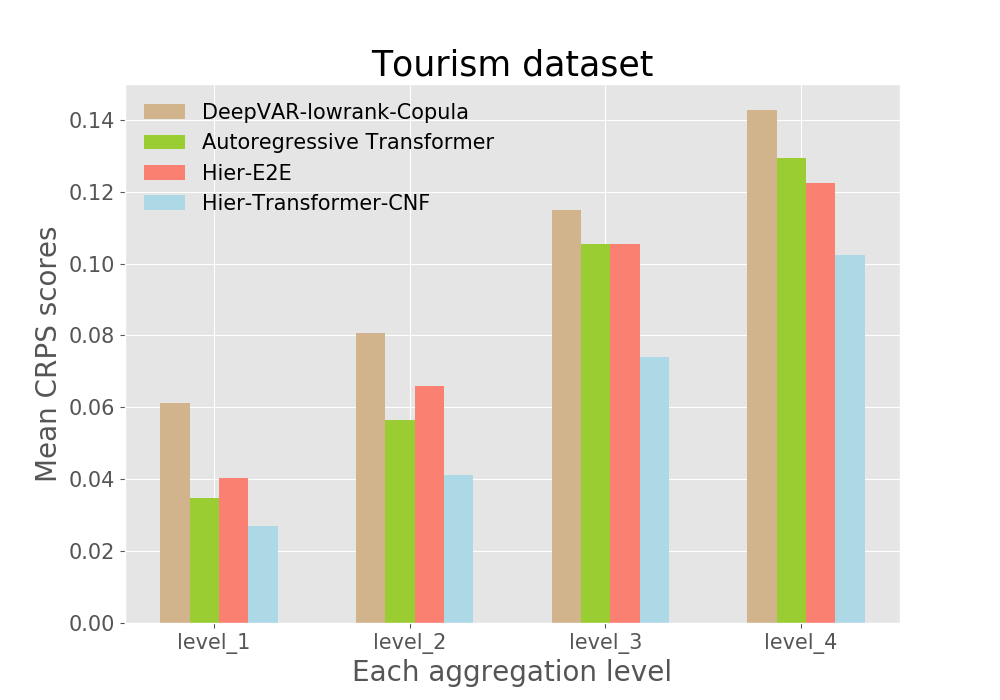}
\end{figure}

\textbf{DeepVAR-lowrank-copula} is an RNN-based model using the lowrank Gaussian coupla \cite{salinas2019high} and \textbf{Autoregressive Transformer} is based the encoder-decoder transformer architecture \cite{katharopoulos2020transformers}, enabled by its multi-head self-attention mechanism to capture both long- and short-term dependencies in time series data. \textbf{Hier-E2E} is the end-to-end model combing DeepVAR(Gaussian distribution) and projection matrix \cite{rangapuram2021end} 

It is obvious that transformer-based models outperform RNN-based models, especially in the upper-level data. The transformer allows the access to any part of the historic data regardless of temporal distance and is thus capable of generating better conditioning for NF head, evidenced by our experiments. 

The non-Gaussian data distribution and nonlinear correlations attest the need for more expressive density estimators, i.e., CNF, which makes our approach significantly outperforming other methods especially in the bottom-level data.  Moreover, the end-to-end reconciliation method also beats direct prediction model in the forecasting at all levels. 

In summary, significant improvements at all levels prove the the effectiveness of the our approach.


\section{Conclusion}
In this paper, we proposed a novel end-to-end approach that tackles forecasting and reconciliation simultaneously for hierarchical time series, without requiring any explicit post-processing step, by combining \textit{autoregressive transformer} and \textit{conditioned normalizing flow} to generate the coherent probabilistic forecasts. 
We conducted extensive empirical evaluations on real-world datasets and demonstrated the competitiveness of our method under various conditions against other state-of-the-art methods. Our ablation study proved the efficacy of every design component we employed.  We have also successfully deployed our method in Alipay as continuous and robust forecasting services for server traffic prediction to promote energy efficiency.

\bibliographystyle{IEEEtran}
\bibliography{sample-base}


\end{document}